\pdfoutput=1

\documentclass[11pt]{article}

\usepackage{ACL2023}

\usepackage{times}
\usepackage{latexsym}

\usepackage{bbding}
\usepackage{amsmath}
\usepackage{amssymb}
\usepackage{arydshln}
\usepackage{bbm}
\usepackage{booktabs}
\usepackage{CJKutf8}
\usepackage{combelow}
\usepackage{enumitem}
\usepackage{epsfig}
\usepackage{fancybox}
\usepackage{float}
\usepackage{graphicx}
\usepackage{hyperref}
\usepackage{multirow}
\usepackage{multicol}
\usepackage{pgfplots}
\usepackage{pifont}
\usepackage{setspace}
\usepackage{subfigure}
\usepackage{tabularx}
\usepackage{url}
\usepackage{xspace}
\usepackage[vlined,ruled,commentsnumbered,linesnumbered]{algorithm2e}

\usepackage{epigraph} 

\usepackage[T1]{fontenc}

\usepackage[utf8]{inputenc}

\usepackage{microtype}

\usepackage{inconsolata}

\usepackage{scalerel,xparse}

\title{EmojiLM: Modeling the New Emoji Language}

\author{Letian Peng, Zilong Wang, Hang Liu, Zihan Wang, Jingbo Shang\thanks{$\ $  Corresponding author. } \\
University of California, San Diego \\
  \texttt{\{lepeng, ziw049, hal064, ziw224, jshang\}@ucsd.edu}
  }

\newcommand{\dataset}{Text2Emoji\xspace}
\newcommand{\model}{EmojiLM\xspace}

\begin{document}
\maketitle
\begin{abstract}
With the rapid development of the internet, online social media welcomes people with different backgrounds through its diverse content. The increasing usage of emoji becomes a noticeable trend thanks to emoji's rich information beyond cultural or linguistic borders. However, the current study on emojis is limited to single emoji prediction and there are limited data resources available for further study of the interesting linguistic phenomenon. To this end, we synthesize a large text-emoji parallel corpus, \dataset, from a large language model. Based on the parallel corpus, we distill a sequence-to-sequence model, \model, which is specialized in the text-emoji bidirectional translation. Extensive experiments on public benchmarks and human evaluation demonstrate that our proposed model outperforms strong baselines and the parallel corpus benefits emoji-related downstream tasks\footnote{Video: \href{https://youtu.be/wwTbbDg2QHM}{youtu.be/wwTbbDg2QHM}}\footnote{Our code is released at \href{https://github.com/KomeijiForce/EmojiLM}{KomeijiForce/EmojiLM}}.

\end{abstract}

\section{Introduction}


\label{sec:intro}

These years have witnessed the boom of social media. Online messages, posts, and articles play indispensable roles in everyone's daily life. Compared with traditional media, social media nowadays provide a much more diverse platform for people all over the world. One noticeable improvement is the various content formats. The emergence of emoji is revolutionary but also foreseeable given their absolute necessity and rich information beyond the cultural or linguistic borders.

Emojis are a pre-defined set of pictograms, logograms, ideograms, or smileys and are usually embedded in electronic messages and web pages to fill in emotional cues otherwise missing from contexts\footnote{\href{https://www.wikipedia.org/wiki/Emoji}{www.wikipedia.org/wiki/Emoji}}. With the increasing use of emojis, they are no longer a ``nice-to-have'' for an existing language, but instead they act as an emerging ``new language'' that can be easily understood by users with diverse backgrounds. As reported, more and more people replace their redundant and bland pure-text posts with vivid emojis to deliver their complex semantics in a more witty way\footnote{\href{https://blog.emojipedia.org/top-emoji-trends-of-2021/}{blog.emojipedia.org/top-emoji-trends-of-2021/}}.

\begin{figure}
    \centering
    \includegraphics[width=\linewidth]{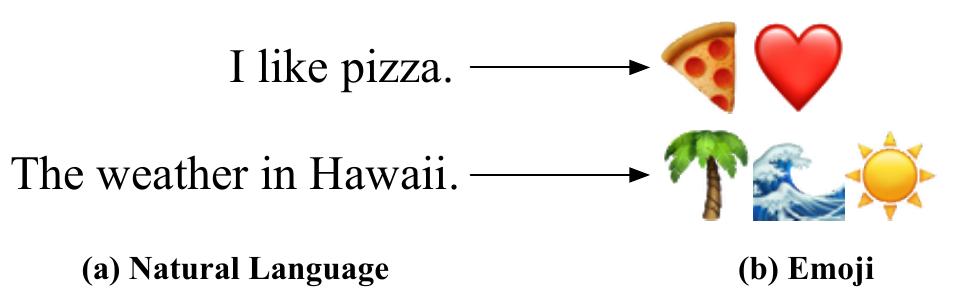}
    \caption{Examples of emoji translation from our Enlgish-Emoji parallel corpus, \dataset.}
    \label{fig:emoji-translation-example}
\end{figure}

Considering the fascinating features of emojis, the existing emoji study is scarce and limited to single emoji predictions~\cite{barbieri2018semeval,lee2022multiemo,singh2022emoji} which focuses on predicting a single emoji character according to the input sentence. While such experiment setting inspires the following research on the new type of data, it underestimates the expressing ability of emojis. Emojis are more than a substitute for emotion words (\texttt{happy}, \texttt{sad}, \texttt{angry}), but they can perform a similar role as natural language when people compose a sentence with multiple emojis, as shown in Figure \ref{fig:emoji-translation-example}. 
Little existing study pay enough attention to the formulation of the multi-emoji experiment setting.

\begin{figure}
    \centering
    \includegraphics[width=\linewidth]{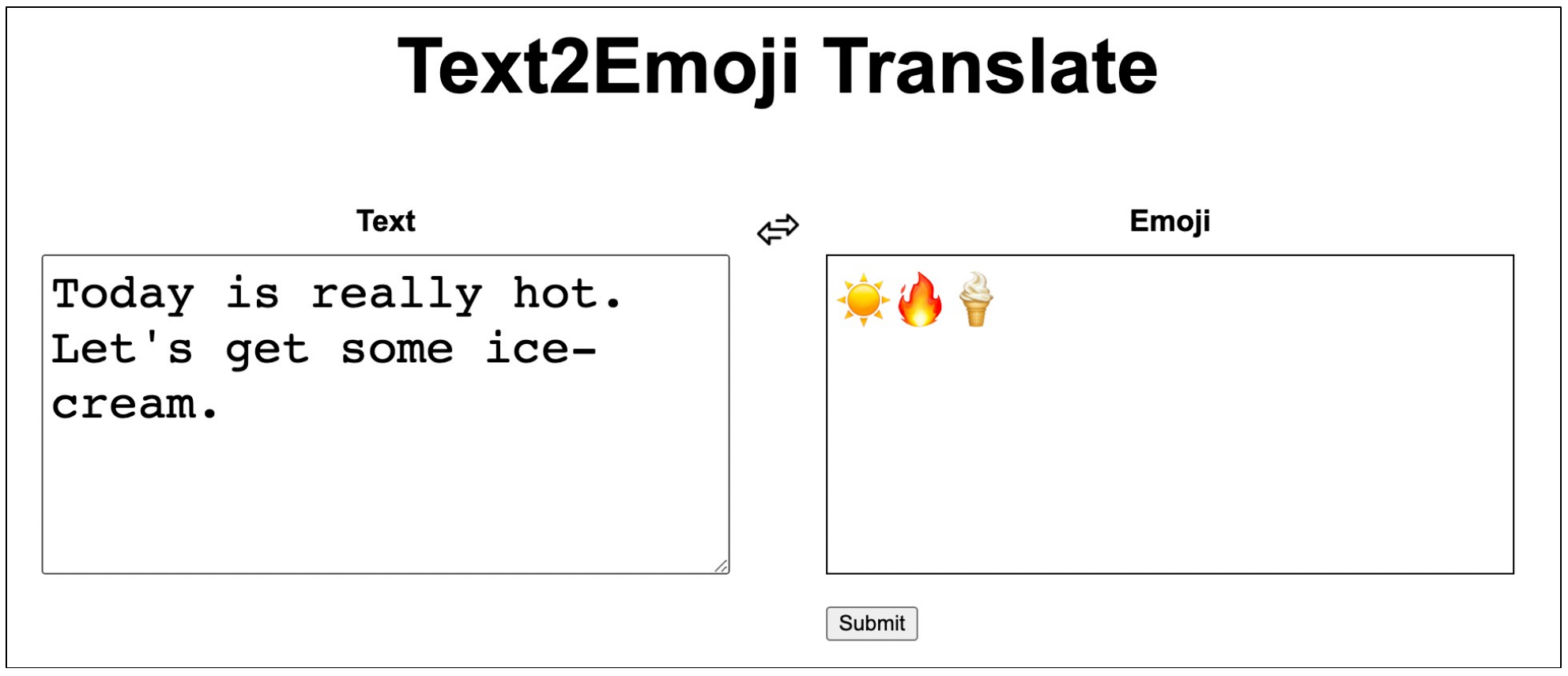}
    \caption{Screenshot from our implemented Text2Emoji translation website.}
    \label{fig:website-screenshot}
\end{figure}

To this end, we propose \dataset, the first parallel corpus for emoji and text, and, \model, a distilled language model specialized in bidirectional English-Emoji translation. To the best of our knowledge, this is the first practice to go beyond rule-based or heuristic ways (e.g. string matching) and bring in the notion of Language Emoji translation for emoji studies. In addition, as shown in Figure \ref{fig:website-screenshot}, we build a website and implement a Chrome extension for users to use our service. They can enjoy the fun and convenience of English-Emoji translation as simply by visiting Google Translate. In comparison with directly querying online large language models services (e.g. ChatGPT\footnote{\href{https://chat.openai.com/}{chat.openai.com/}}, Bard\footnote{\href{https://bard.google.com/}{bard.google.com/}}), our model is lighter and cheaper with weights accessible to the public community. Moreover, it is also feasible to fine-tune our proposed \model for relevant downstream tasks, including the aforementioned single-emoji prediction task.

To build the Text-Emoji parallel corpus, we synthesize data from large language models, which allows us to effortlessly create a corpus containing emojis that have 100 times larger emoji vocabulary size than prior work, such as TweetEval~\cite{tweeteval}. 
We spend most of our effort on automatic and manual evaluation of the quality of the corpus and the derived distilled model.
As aforementioned, there are little existing annotated data and it is infeasible to ask human laborers for annotation because of the high cost. We design prompts to trigger large language models to generate the corresponding emoji sequence according to the provided sentence. Next, similar to previous practices in neural machine translation, we train a sequence-to-sequence model on the parallel data as an English-Emoji translator. We package the pre-trained model and the I/O interfaces into a website and a Chrome extension for the public to use. To evaluate the performance of \model, we also compare it with strong baselines on three popular benchmarks, TweetEval, AG-News, and DBPdeia. Extensive experiments demonstrate the effectiveness of our proposed approach. Moreover, we conduct human evaluation on \dataset and \model, where we compare our translation results with the ground truth in our parallel corpus. We observe that human annotators cannot distinguish the translation results from the ground truth, showing that our proposed model serves as an emoji translator of high quality.

We summarize our contribution as follows.
\begin{itemize}
    \item We propose a parallel corpus for English-Emoji translation, \dataset, which makes it possible to study the emoji usage as a new form of language and extends the scope of current emoji study from the single-Emoji prediction to the emoji translation.
    \item Based on \dataset, we distill a sequence-to-sequence model specialized in bidirectional English-Emoji translation, and implement a website and a chrome extension for public to use.
    \item Extensive experiments on public benchmark and human evaluation demonstrate the effectiveness of \model on three representative benchmarks.
\end{itemize}

\section{Related Work}
\paragraph{Research on Emoji}

As mentioned in Section \ref{sec:intro}, the current research on emojis mostly treats them as a symbol of emotion and creates the experiment setting, the single emoji prediction, which asks models to classify a given sentence into a given set of emojis. SemEval 2018~\cite{barbieri2018semeval} proposes a multilingual emoji prediction task. The participants are asked to predict a single emoji for the given tweets either in English or Spanish. The emojis come from a fixed set of top 20 most popular emojis of each language. MultiEmo~\cite{lee2022multiemo} proposes to use Bi-LSTM and attention mechanism to better solve the task. \citet{singh2022emoji} leverages the multi-task training and trains the model on emotion detection, sentiment analysis, and emoji prediction together. 

Another research trend on emoji focuses on hate detection in online text with emojis~\cite{kirk2021hatemoji,das2023evaluating}. These works argue that existing methods fail to pay enough attention to the emoji characters in the online text, leading to their incapability to detect hateful language. While these works on hate detection emphasize the importance of emoji in online text, they are limited by the limited amount of existing online tweets or posts with emojis. Our work builds up a comprehensive parallel corpus for emoji so as to facilitate more future research on the emoji study.

\paragraph{Data Synthesis with LLMs}
Data synthesis is widely used to generate training data when it is difficult to obtain enough data for a specific task. \citet{tang2023does,hamalainen2023evaluating} leverage LLMs to generate synthetic data for the medical or HCI domains, and the synthesized data bring about improvement to the task thanks to the outstanding ability of LLMs.

\section{EmojiLM}
\subsection{English-Emoji Parallel Corpus}

We build a large English-Emoji parallel corpus, \dataset, by prompting the LLM, \texttt{gpt-3.5-turbo} \cite{chatgpt}. To fully cover emojis in different domains, we ask the LLM to propose a sentence given a certain domain and then translate it into a series of emojis. We include $19$ domains: ``feeling'', ``career'', ``clothes'', ``animal'', ``plant'', ``weather'', ``food'', ``sports'', ``arts'', ``vehicle'', ``building'', ``tool'', ``country'', ``electrical appliance'', ``activity'', ``experience'', ``family member''. We select these domains according to the categorization of emojis. Our prompt is presented as follows,

\begin{center}
    \textit{Write some sentences about a kind of <topic> and their pure emoji series translations in the following format: Text:... Emoji Translation:...}
\end{center}

For the startup, we query the LLM $1000$ times for each topic and collect the generated parallel sentences. We increase the diversity by setting the temperature of the generation to $1.5$. However, the diversity is still limited since the sampling follows the probability distribution of a constant prompt. Thus, we utilize a property of the LLM to further increase the diversity that LLM inclines to generate something different given an instance for the instruction. Thus, with the startup data, we combine the prompt with an instance to encourage the LLM to generate more diverse sentences inside the topic. 

\begin{center}
    User: \textit{Write some sentences about a kind of <topic> and their pure emoji series translations in the following format: Text:... Emoji Translation:...}
    System: \textit{Text: <text> Emoji: <emoji>}
\end{center}

\noindent where we prompt the LLM to generate after a parallel instance, which is randomly sampled from the pool of generated instances. We further prompt with instances for $15000$ times and thus build a large English-Emoji parallel corpus with $503.7K$ instances by filtering non-emoji tokens in the emoji series and instances with no emoji. 

\begin{table}
\centering
\scalebox{1.}{
\begin{tabular}{lc}
\toprule
Attribute & Value \\
\midrule
\#Instance & $503.7K$ \\
\#Emoji Vocabulary & $2.3K$ \\
\#Text Average Length & $15.18$ \\
\#Emoji Average Length & $7.97$ \\
\bottomrule
\end{tabular}
}
\caption{The statistics of our English-Emoji parallel.} 
\label{tab:stat}
\end{table}

\begin{figure}
    \centering
    \includegraphics[width=0.99\linewidth]{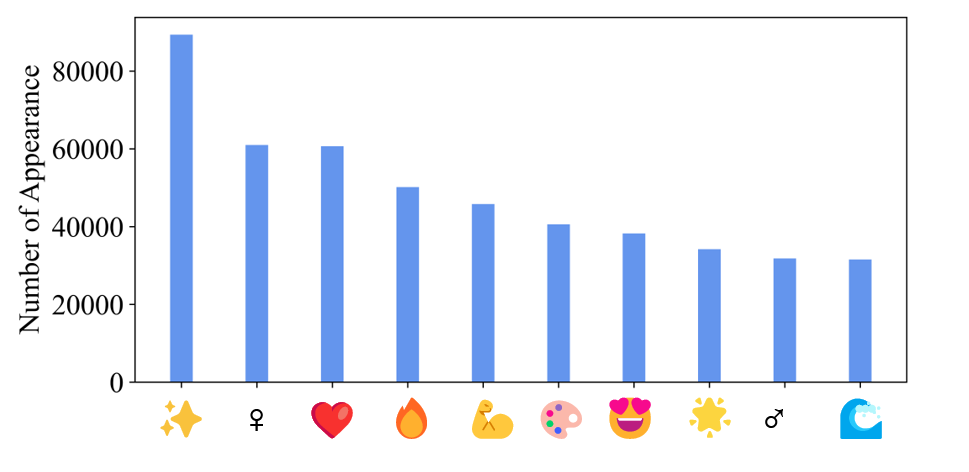}
    \caption{Emoji tokens with the top $10$ appearance in \dataset.}
    \label{fig:stat_emoji}
\end{figure}

\paragraph{Dataset Statistics} In Table~\ref{tab:stat}, we present the statistics of our English-Emoji parallel corpus. Our corpus covers a large ($2.3K$) emoji vocabulary, which is more than $100$ times larger than the emoji class number ($20$) of the popular single emoji prediction dataset, TweetEval \cite{tweeteval}. The texts are in medium lengths, which is similar to the application scenario of emojis in reality. We also show the most popular emojis in Figure~\ref{fig:stat_emoji}, which are also most commonly used in real-life social media. The gender symbols frequently appear in our corpus because there are composed emojis that are encoded as the combination of multiple emojis, joined with ``$\backslash$u200d''. Gender symbols are generally combined with human emojis to switch the gender of them. 

\subsection{Learning Bidirectional Translation}

Utilizing our comprehensive English-Emoji parallel corpus, we embark on the journey of training models that can seamlessly translate between texts and emojis in both directions. We adopted the encoder-decoder framework, a popular architecture for translation tasks, and implemented it with models like BART \cite{BART} and T5 \cite{T5}.

However, it's important to note that emojis present unique tokenization challenges. The pre-existing vocabularies in models such as BART and T5 aren't optimized for emojis. As such, we enhanced the tokenizers by integrating the new emoji vocabulary, ensuring they recognize and effectively process these symbols.

A distinct challenge in emoji translation is the concept of composed emojis. These emojis are encoded as a sequence, with different symbols joined using the ``$\backslash$u200d'' separator. For instance, certain human emojis can switch genders by being combined with gender symbols. To ensure our translator is attuned to this nuance, we tokenize composed emojis into their constituent parts, separating them with the ``$\backslash$u200d'' token. This strategy helps our translation models grasp the intricate relationships and compositions that emojis can present.

\section{Experiment}
\subsection{Emoji Language Modeling}

For both translation directions, we train a BART-Base (BART-B) model for $2$ epochs with a $32$ batch size. The learning rate is set to $5\times 10^{-5}$. We set $2000$ warmup steps for emoji-to-text translation.

We first evaluate the performance of learning on the English-Emoji parallel corpus. In this experiment, we split the whole corpus into train/dev/test datasets by $8$/$1$/$1$. We train different language models on the training split and then translate the test split. In the generation step, we set the beam size to $4$ and temperature to $1.0$ to search for the sequence that maximizes the existence probability. 

\begin{figure}
    \centering
    \includegraphics[width=0.99\linewidth]{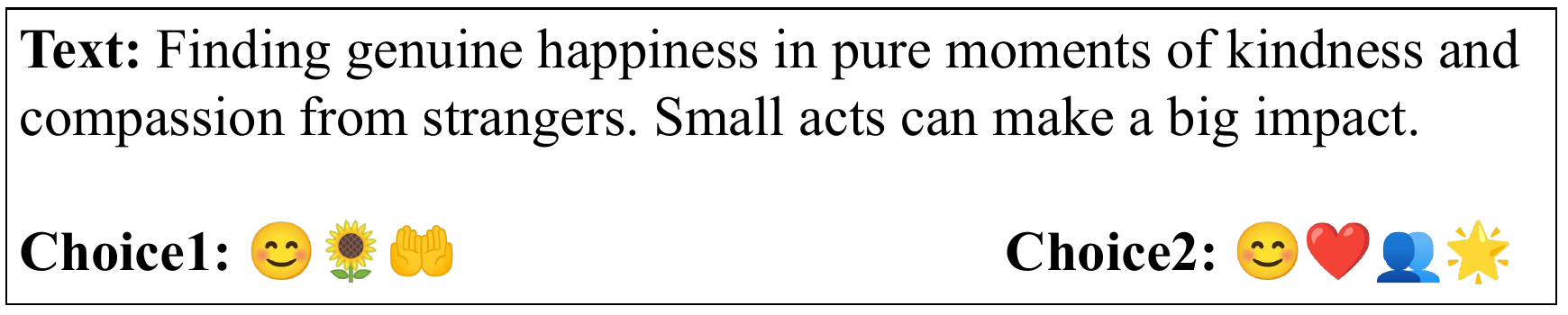}
    \caption{An example in human evaluation.}
    \label{fig:task_example}
\end{figure}

\begin{table}
\centering
\scalebox{.94}{
\begin{tabular}{llccccc}
\toprule
& Model & B1 & B2 & B3 & B4 & BS \\
\midrule
\multirow{3}*{\rotatebox{90}{T$\rightarrow$E}} & BART-B & $32.2$ & $20.0$ & $14.2$ & $10.8$ & - \\
& T5-B & $32.6$ & $20.4$ & $14.5$ & $11.0$ & - \\
& BART-L & $34.8$ & $21.9$ & $16.1$ & $12.0$ & - \\
\midrule
\multirow{3}*{\rotatebox{90}{E$\rightarrow$T}} & BART-B & $24.9$ & $16.8$ & $12.8$ & $10.3$ & $36.5$ \\
& T5-B & $25.3$ & $17.0$ & $12.9$ & $10.3$ & $36.3$ \\
& BART-L & $25.8$ & $17.4$ & $13.2$ & $10.6$ & $36.5$  \\
\bottomrule
\end{tabular}
}
\caption{The performance of bidirectional translation on the English-Emoji parallel corpus. (T5-B has the same scale as BART-L.)} 
\label{tab:emojilm}
\end{table}

\begin{table}
\centering
\scalebox{.83}{
\begin{tabular}{llcccc}
\toprule
\multicolumn{2}{c}{Dataset} & \#Train & \#Test & \#Domain & \#Label \\
\midrule
\multirow{4}*{\rotatebox{90}{TweetEval}} & Emoji & $45.0K$ & $50.0K$ & Emoji & $20$ \\
& Emoji-EX & $47.4K$ & $5.3K$ & Emoji & $32$ \\
& Sentiment & $45.6K$ & $12.3K$ & Emotion & $3$ \\
& Emotion & $3.3K$ & $1.4K$ & Emotion & $4$ \\
\multicolumn{2}{l}{AG\_News} & $120.0K$ & $7.6K$ & Topic & $4$ \\
\multicolumn{2}{l}{DBPedia} & $560.0K$ & $70.0K$ & Topic & $14$ \\
\bottomrule
\end{tabular}
}
\caption{The statistics of datasets used for task transferring.} 
\label{tab:test_stat}
\end{table}

\begin{table*}
\centering
\scalebox{0.9}{
\begin{tabular}{llcccccc}
\toprule
& \multirow{2}*{Dataset} & \multicolumn{4}{c}{TweetEval} & \multirow{2}*{AG-News} & \multirow{2}*{DBPedia} \\
& & Emoji & Emoji-EX & Sentiment & Emotion \\
& \#Label & $20$ & $32$ & $3$ & $4$ & $4$ & $14$ \\
\midrule
\multirow{4}*{\rotatebox{90}{\textsc{Full Sup.}}}& BERT & $31.3$ & $8.1$ & $67.1$ & $79.4$ & $94.0$ & $98.8$ \\
& BERTweet & $33.6$ & $10.1$ & $70.1$ & $79.9$ & $\textbf{94.3}$ & $98.9$ \\
& BART & $30.8$ & $12.1$ & $69.8$ & $79.8$ & $93.4$ & $\textbf{99.0}$ \\
& EmojiLM & $\textbf{34.8}$ & $\textbf{23.5}$ & $\textbf{70.4}$ & $\textbf{81.3}$ & $93.6$ & $\textbf{99.0}$ \\
\midrule
\multirow{4}*{\rotatebox{90}{\textsc{Few Sup.}}}& BERT & $10.3$ & $5.1$ & $36.3$ & $27.0$ & $54.9$ & $95.1$ \\
& BERTweet & $15.4$ & $8.8$ & $33.8$ & $35.1$ & $69.4$ & $94.6$ \\
& BART & $11.4$ & $10.4$ & $51.6$ & $58.3$ & $75.0$ & $95.8$ \\
& EmojiLM & $\textbf{23.8}$ & $\textbf{13.6}$ & $\textbf{55.7}$ & $\textbf{61.3}$ & $\textbf{84.0}$ & $\textbf{96.4}$ \\
\bottomrule
\end{tabular}
}
\caption{Task transferring performances of different pre-trained language models.} 
\label{tab:transfer}
\end{table*}

\paragraph{Human Evaluation} To gain insight into how humans perceive the quality of our translation results, we engage human evaluators to assess our parallel corpus. Recognizing that directly judging the correlation between a series of emojis and the corresponding text can be challenging and subjective, we adopt an indirect method of evaluation. Specifically, we present the evaluators with two potential translation outcomes, asking them to choose the one that seems more appropriate as shown in Figure~\ref{fig:task_example}. To eliminate biases linked to the order of presentation, the options are randomly switched. In this evaluation process, we sample $200$ instances, with each instance being assessed by $3$ different evaluators. This task is divided among $15$ human evaluators, meaning that each evaluator provides feedback on $40$ samples. After compiling the input from this substantial number of evaluators, we determine the final result by selecting the choice that has been favored by the majority.

1) \textbf{Text-to-emoji evaluation:} The human evaluator is given a sentence and is asked to select the better emoji series that represents the text between the generated emoji series by EmojiLM and one directly from the corpus. For $40\%$ of the sample texts, participants favor the translation from our EmojiLM. This high proportion demonstrates that EmojiLM has a fair performance compared with the emoji from the corpus. 

2) \textbf{Emoji-to-text evaluation:} This is a mirror test to the text-to-emoji evaluation. Given the emoji, the human evaluator selects the better-translated text from the emoji series. For $45\%$ of the emojis, text from the corpus is selected as a better translation, further verifying the emoji-to-text ability of EmojiLM. 

3) \textbf{Comparison w/ string-matching translator:} The evaluator is asked to compare the emoji generated by EmojiLM and an existing Emoji-Translate python package\footnote{\href{https://pypi.org/project/emoji-translate}{pypi.org/project/emoji-translate}}, which uses the string-matching strategy to map words into corresponding emojis. From 200 sample texts, EmojiLM is selected as a better emoji translator for $88\%$ of them, verifying the absolute advantage of our translator.

Overall, both emoji and text generated by EmojiLM hold a high level of quality in representing their respective input. EmojiLM consistently produces a higher quality of emojis when compared to the emojis generated by the existing Emoji-Translate python package.

\paragraph{Automatic Evaluation} We involve more language models on automatic evaluation, including larger BART (BART-L) and T5 (T5-B, with the same parameter scale as BART-L). The learning rate is set to $3\times 10^{-4}$ for T5. The instruction for T5 is ``translate text (emoji series) into emoji series (text): ''. We use BLUE-$n$ (B$n$) \cite{bleu} and the BERTScore (BS, only for emoji-to-text evaluation) metric \cite{bertscore} to automatically evaluate their performance on the test dataset. The translation performances are shown in Table~\ref{tab:emojilm}. The bidirectional translator achieves high accuracy in emoji series generation considering the high diversity of emoji usage. For the emoji-to-text translation, the performance on BERTScore suggests our translator is able to capture the most semantics in the input emoji series. 

\subsection{Task Transferring}

\begin{figure}
    \centering
    \includegraphics[width=0.99\linewidth]{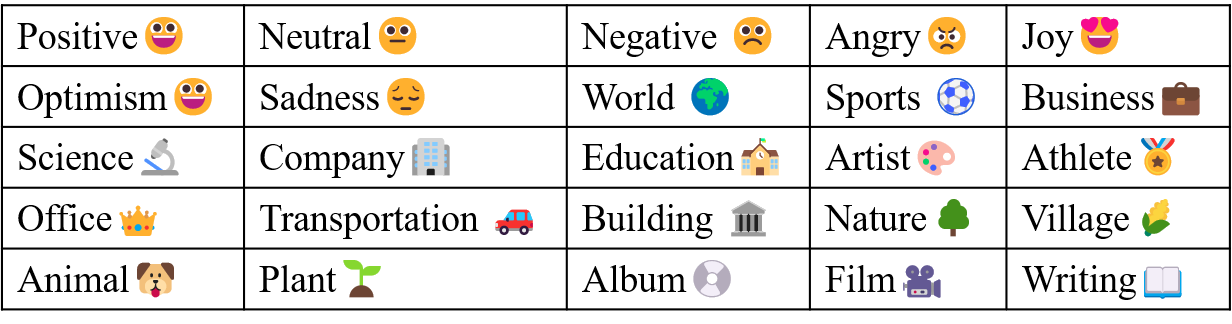}
    \caption{The emojis used as labels for task transferring.}
    \label{fig:emoji_labels}
\end{figure}

\paragraph{Datasets and Metrics} We explore the transferability of emoji modeling to other tasks that are relevant to emojis. \textbf{TweetEval} \cite{tweeteval} includes a popular subset (\textbf{Emoji}) for single emoji prediction with $20$ emojis as labels. \textbf{Emoji-EX}\footnote{\href{https://huggingface.co/datasets/adorkin/extended_tweet_emojis/viewer/adorkin--extended_tweet_emojis/}{huggingface.co/datasets/adorkin/extended\_tweet\_emojis/\\viewer/adorkin--extended\_tweet\_emojis}} is a dataset that extends the emoji number to $32$ by incorporating negative emojis. \textbf{Emotion} ($6$ classes) and \textbf{Sentiment} ($3$ classes) are two subsets under \textbf{TweetEval} that can be formalized as emoji prediction by presenting the labels as emojis. To verify the generality of task transferring, we further incorporate two topic classification tasks: \textbf{AG-News} ($4$ classes) \cite{ag} and \textbf{DBPedia} ($14$ classes) \cite{dbpedia}, also by presenting label names as emojis. We show the emojis formalization in Figure~\ref{fig:emoji_labels}. More specific statistics of datasets in our experiments are shown in Table~\ref{tab:test_stat}. Following the previous work \cite{tweeteval}, we use macro F1 score to evaluate performances on those text classification tasks.

\paragraph{Baselines and Setups} As BART performs much better than T5 on translating text to emojis, we select BART for transferring to relevant tasks. We include different pre-trained language models as the baselines such as BERT \cite{DBLP:conf/naacl/DevlinCLT19} and its variant, BERTweet \cite{bertweet}, which is designed to be pre-trained on tweet texts. We also include BART without training on the English-Emoji parallel corpus to verify the benefit of further learning. For the BART baseline, we predict label texts instead of emojis, which is more consistent with the pre-training process of BART. We use models with the \texttt{large} size in the experiments for a fair comparison. For experiments in this part, we use a text-to-emoji translator that is pre-trained on the full corpus. We train models for $5$ epochs for datasets without validation splits. All results in our experiments are averaged by $5$ runs. 

The model performances on task transferring are presented in Table~\ref{tab:transfer}. Our model shows the most significant advantage in emoji predictions, which outperforms all baselines, including the strong BERTweet. There is a definite gap between our EmojiLM and other baselines when the emoji classes increase to $32$ in the Emoji-EX dataset. We attribute this to the understanding of emoji semantics of our EmojiLM, which enables it to handle the nuance between similar emojis. Our text-to-emoji pre-training also benefits emotion prediction as EmojiLM also achieves the best performances on Sentiment and Emotion subsets. This is also consistent with the fact that emojis are able to represent emotions efficiently, and thus the pre-training on emoji modeling benefits the emotion prediction task. On topic classification tasks, there is not a significant improvement under full supervision perhaps due to the large scale of those training splits. In comparison to the BART baseline, our EmojiLM always achieves a better performance, which suggests the potential of our model when labels can be formalized as emojis in text classification tasks.

\paragraph{Low-resource Situation} Another advantage of training on our English-Emoji parallel corpus is improving the few-shot task transferring ability. As emoji prediction can be viewed as text-to-one-emoji translation, the consistency between the upstream and downstream tasks supports the potential for an efficient few-shot learning scenario. In Table~\ref{tab:transfer}, we present the performances of different language models using $n$-way $10$-shot data as the supervision. We observe a large gap between our EmojiLM and other baselines. This phenomenon can be attributed to two factors: the translation scenario and the emoji understanding. The comparison between BART and BERT (BERTweet) reveals the first factor as BART performs much better with few data. The label names play an important role in task understanding when the supervision from the data scale decreases. Another factor is the concise nature of the emoji that summarizes a topic with a single token (emoji). Thus, our text-to-emoji EmojiLM significantly outperforms the text-to-text BART baseline. 

\subsection{Case Study}

\begin{figure}
    \centering
    \includegraphics[width=0.99\linewidth]{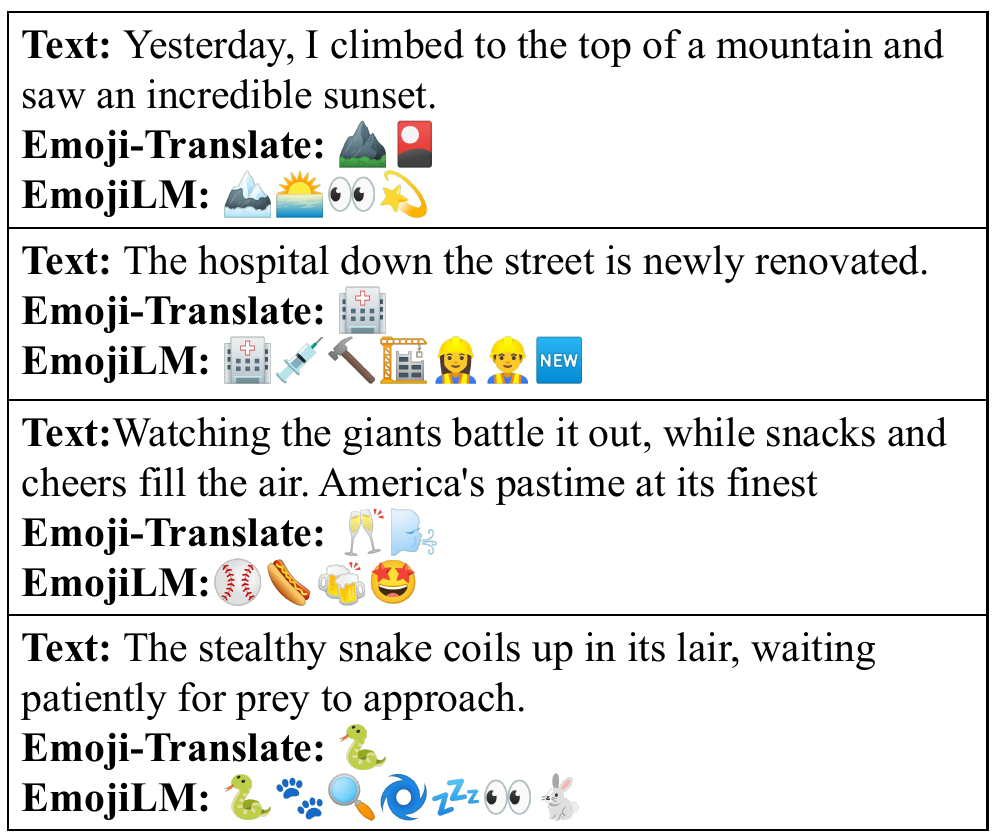}
    \caption{Case study by comparing our translator (EmojiLM) with the Emoji-Translate package.}
    \label{fig:case}
\end{figure}


In Figure~\ref{fig:case}, we use specific cases to compare our EmojiLM translation results with those from the Emoji-Translate package, a string-matching baseline. The first two examples reveal a richer knowledge base of EmojiLM, as it accurately detects phrases like “incredible sunset” and “newly renovated,” translating them into corresponding emojis. Unlike Emoji-Translate, EmojiLM also recognizes specific contexts. In the third example, it identifies “the Giants” as a baseball team, not literal giants, and outputs the emojis of relevant snacks such as “hotdog” and “beer.”

In the fourth example, the power of our EmojiLM to create a narrative emoji series is clear. It crafts a story where a snake is patiently hunting, while Emoji-Translate merely returns a single snake emoji. This highlights the limitation of  Emoji-Translate: it detects individual words and returns corresponding emojis but often misses the actual meaning within a specific context. On the other hand, EmojiLM offers a nuanced understanding of both the words and the overall sentence, utilizing a broader emoji knowledge base to provide more accurate and engaging translations.



\section{Conclusion and Future Work}

This paper introduces EmojiLM, an English-Emoji translator that goes beyond single emoji prediction to handle complex translations between texts and emojis. Utilizing a large corpus, EmojiLM outperforms existing baselines like BERTweet in text classification. This work represents a significant advancement in understanding and leveraging emojis in linguistic phenomena. Future work will delve into the intricate understanding of emojis by employing the embeddings derived from our model. Furthermore, we will explore the multifaceted association between textual and visual characteristics, with emojis serving as a connecting bridge. 

\section*{Limitation}

While the work presented herein demonstrates promising strides in the field of Emoji translation, certain limitations must be acknowledged. First and foremost, although we have successfully utilized LLMs to construct an English-Emoji parallel corpus, the source of this ability remains somewhat enigmatic. There is no publicly available training data specifically targeting this task, leading us to hypothesize that this ability stems from the LLM's understanding of emoji semantics and translation tasks. However, the exact mechanism remains unexplored and warrants further investigation. Secondly, the corpus itself might harbor biases towards popular emojis, reflecting the corpus used to train the LLMs rather than a balanced representation of Emoji usage across different cultures and communication contexts. This bias could lead to an overemphasis on commonly used emojis and potentially overlook the nuanced usage of lesser-known emojis, thus limiting the generalization and applicability of the derived models and tools. These challenges underline areas for future research and refinement in the pursuit of a more robust and comprehensive understanding of Emoji as a form of language.



\bibliography{anthology,custom}

\begin{thebibliography}{17}
\expandafter\ifx\csname natexlab\endcsname\relax\def\natexlab#1{#1}\fi

\bibitem[{Auer et~al.(2007)Auer, Bizer, Kobilarov, Lehmann, Cyganiak, and Ives}]{dbpedia}
S{\"{o}}ren Auer, Christian Bizer, Georgi Kobilarov, Jens Lehmann, Richard Cyganiak, and Zachary~G. Ives. 2007.
\newblock \href {https://doi.org/10.1007/978-3-540-76298-0\_52} {Dbpedia: {A} nucleus for a web of open data}.
\newblock In \emph{The Semantic Web, 6th International Semantic Web Conference, 2nd Asian Semantic Web Conference, {ISWC} 2007 + {ASWC} 2007, Busan, Korea, November 11-15, 2007}, volume 4825 of \emph{Lecture Notes in Computer Science}, pages 722--735. Springer.

\bibitem[{Barbieri et~al.(2020)Barbieri, Camacho{-}Collados, Anke, and Neves}]{tweeteval}
Francesco Barbieri, Jos{\'{e}} Camacho{-}Collados, Luis~Espinosa Anke, and Leonardo Neves. 2020.
\newblock \href {https://doi.org/10.18653/v1/2020.findings-emnlp.148} {Tweeteval: Unified benchmark and comparative evaluation for tweet classification}.
\newblock In \emph{Findings of the Association for Computational Linguistics: {EMNLP} 2020, Online Event, 16-20 November 2020}, volume {EMNLP} 2020 of \emph{Findings of {ACL}}, pages 1644--1650. Association for Computational Linguistics.

\bibitem[{Barbieri et~al.(2018)Barbieri, Camacho-Collados, Ronzano, Anke, Ballesteros, Basile, Patti, and Saggion}]{barbieri2018semeval}
Francesco Barbieri, Jose Camacho-Collados, Francesco Ronzano, Luis~Espinosa Anke, Miguel Ballesteros, Valerio Basile, Viviana Patti, and Horacio Saggion. 2018.
\newblock Semeval 2018 task 2: Multilingual emoji prediction.
\newblock In \emph{Proceedings of the 12th international workshop on semantic evaluation}, pages 24--33.

\bibitem[{Das et~al.(2023)Das, Pandey, and Mukherjee}]{das2023evaluating}
Mithun Das, Saurabh~Kumar Pandey, and Animesh Mukherjee. 2023.
\newblock Evaluating chatgpt's performance for multilingual and emoji-based hate speech detection.
\newblock \emph{arXiv preprint arXiv:2305.13276}.

\bibitem[{Devlin et~al.(2019)Devlin, Chang, Lee, and Toutanova}]{DBLP:conf/naacl/DevlinCLT19}
Jacob Devlin, Ming{-}Wei Chang, Kenton Lee, and Kristina Toutanova. 2019.
\newblock \href {https://doi.org/10.18653/v1/n19-1423} {{BERT:} pre-training of deep bidirectional transformers for language understanding}.
\newblock In \emph{Proceedings of the 2019 Conference of the North American Chapter of the Association for Computational Linguistics: Human Language Technologies, {NAACL-HLT} 2019, Minneapolis, MN, USA, June 2-7, 2019, Volume 1 (Long and Short Papers)}, pages 4171--4186. Association for Computational Linguistics.

\bibitem[{H{\"a}m{\"a}l{\"a}inen et~al.(2023)H{\"a}m{\"a}l{\"a}inen, Tavast, and Kunnari}]{hamalainen2023evaluating}
Perttu H{\"a}m{\"a}l{\"a}inen, Mikke Tavast, and Anton Kunnari. 2023.
\newblock Evaluating large language models in generating synthetic hci research data: a case study.
\newblock In \emph{Proceedings of the 2023 CHI Conference on Human Factors in Computing Systems}, pages 1--19.

\bibitem[{Kirk et~al.(2021)Kirk, Vidgen, R{\"o}ttger, Thrush, and Hale}]{kirk2021hatemoji}
Hannah~Rose Kirk, Bertram Vidgen, Paul R{\"o}ttger, Tristan Thrush, and Scott~A Hale. 2021.
\newblock Hatemoji: A test suite and adversarially-generated dataset for benchmarking and detecting emoji-based hate.
\newblock \emph{arXiv preprint arXiv:2108.05921}.

\bibitem[{Lee et~al.(2022)Lee, Jeong, and Park}]{lee2022multiemo}
SangEun Lee, Dahye Jeong, and Eunil Park. 2022.
\newblock Multiemo: Multi-task framework for emoji prediction.
\newblock \emph{Knowledge-Based Systems}, 242:108437.

\bibitem[{Lewis et~al.(2020)Lewis, Liu, Goyal, Ghazvininejad, Mohamed, Levy, Stoyanov, and Zettlemoyer}]{BART}
Mike Lewis, Yinhan Liu, Naman Goyal, Marjan Ghazvininejad, Abdelrahman Mohamed, Omer Levy, Veselin Stoyanov, and Luke Zettlemoyer. 2020.
\newblock \href {https://doi.org/10.18653/v1/2020.acl-main.703} {{BART:} denoising sequence-to-sequence pre-training for natural language generation, translation, and comprehension}.
\newblock In \emph{Proceedings of the 58th Annual Meeting of the Association for Computational Linguistics, {ACL} 2020, Online, July 5-10, 2020}, pages 7871--7880. Association for Computational Linguistics.

\bibitem[{Nguyen et~al.(2020)Nguyen, Vu, and Nguyen}]{bertweet}
Dat~Quoc Nguyen, Thanh Vu, and Anh~Tuan Nguyen. 2020.
\newblock \href {https://doi.org/10.18653/v1/2020.emnlp-demos.2} {Bertweet: {A} pre-trained language model for english tweets}.
\newblock In \emph{Proceedings of the 2020 Conference on Empirical Methods in Natural Language Processing: System Demonstrations, {EMNLP} 2020 - Demos, Online, November 16-20, 2020}, pages 9--14. Association for Computational Linguistics.

\bibitem[{OpenAI(2023)}]{chatgpt}
OpenAI. 2023.
\newblock \href {https://doi.org/10.48550/arXiv.2303.08774} {{GPT-4} technical report}.
\newblock \emph{CoRR}, abs/2303.08774.

\bibitem[{Papineni et~al.(2002)Papineni, Roukos, Ward, and Zhu}]{bleu}
Kishore Papineni, Salim Roukos, Todd Ward, and Wei{-}Jing Zhu. 2002.
\newblock \href {https://doi.org/10.3115/1073083.1073135} {Bleu: a method for automatic evaluation of machine translation}.
\newblock In \emph{Proceedings of the 40th Annual Meeting of the Association for Computational Linguistics, July 6-12, 2002, Philadelphia, PA, {USA}}, pages 311--318. {ACL}.

\bibitem[{Raffel et~al.(2020)Raffel, Shazeer, Roberts, Lee, Narang, Matena, Zhou, Li, and Liu}]{T5}
Colin Raffel, Noam Shazeer, Adam Roberts, Katherine Lee, Sharan Narang, Michael Matena, Yanqi Zhou, Wei Li, and Peter~J. Liu. 2020.
\newblock \href {http://jmlr.org/papers/v21/20-074.html} {Exploring the limits of transfer learning with a unified text-to-text transformer}.
\newblock \emph{J. Mach. Learn. Res.}, 21:140:1--140:67.

\bibitem[{Singh et~al.(2022)Singh, Chauhan, Firdaus, Ekbal, and Bhattacharyya}]{singh2022emoji}
Gopendra~Vikram Singh, Dushyant~Singh Chauhan, Mauajama Firdaus, Asif Ekbal, and Pushpak Bhattacharyya. 2022.
\newblock Are emoji, sentiment, and emotion friends? a multi-task learning for emoji, sentiment, and emotion analysis.
\newblock In \emph{Proceedings of the 36th Pacific Asia Conference on Language, Information and Computation}, pages 166--174.

\bibitem[{Tang et~al.(2023)Tang, Han, Jiang, and Hu}]{tang2023does}
Ruixiang Tang, Xiaotian Han, Xiaoqian Jiang, and Xia Hu. 2023.
\newblock Does synthetic data generation of llms help clinical text mining?
\newblock \emph{arXiv preprint arXiv:2303.04360}.

\bibitem[{Zhang et~al.(2020)Zhang, Kishore, Wu, Weinberger, and Artzi}]{bertscore}
Tianyi Zhang, Varsha Kishore, Felix Wu, Kilian~Q. Weinberger, and Yoav Artzi. 2020.
\newblock \href {https://openreview.net/forum?id=SkeHuCVFDr} {Bertscore: Evaluating text generation with {BERT}}.
\newblock In \emph{8th International Conference on Learning Representations, {ICLR} 2020, Addis Ababa, Ethiopia, April 26-30, 2020}. OpenReview.net.

\bibitem[{Zhang et~al.(2015)Zhang, Zhao, and LeCun}]{ag}
Xiang Zhang, Junbo~Jake Zhao, and Yann LeCun. 2015.
\newblock \href {https://proceedings.neurips.cc/paper/2015/hash/250cf8b51c773f3f8dc8b4be867a9a02-Abstract.html} {Character-level convolutional networks for text classification}.
\newblock In \emph{Advances in Neural Information Processing Systems 28: Annual Conference on Neural Information Processing Systems 2015, December 7-12, 2015, Montreal, Quebec, Canada}, pages 649--657.

\end{thebibliography}
\bibliographystyle{acl_natbib}

\clearpage

\appendix

\end{document}